%% file: main.tex
\documentclass[runningheads]{llncs}
\usepackage{graphicx}

\usepackage[ruled,vlined]{algorithm2e}
\usepackage{bm}
\usepackage{subcaption}
\usepackage{bbm}
\usepackage{makecell}
\usepackage{multirow}
\usepackage{cuted}

\usepackage{tikz}
\usepackage{comment}
\usepackage{amsmath,amssymb} 
\usepackage{color}

\usepackage{booktabs}
\usepackage{color}

\usepackage{marvosym}
\usepackage{ifsym}
\usepackage[pagebackref,breaklinks,colorlinks]{hyperref}

\usepackage{eucal}
\usepackage[accsupp]{axessibility}  

\makeatletter
\newcommand{\printfnsymbol}[1]{%
  \textsuperscript{\@fnsymbol{#1}}%
}

\begin{document}
\pagestyle{headings}
\mainmatter
\def\ECCVSubNumber{247}  

\title{StyleSwap: Style-Based Generator Empowers Robust Face Swapping} 

\titlerunning{StyleSwap}
\author{Zhiliang Xu\inst{1}\thanks{Equal contribution. \textsuperscript{\Letter} Corresponding author.}
\and
Hang Zhou\inst{1}\printfnsymbol{1}\textsuperscript{\Letter} \and
Zhibin Hong\inst{1} \and
Ziwei Liu\inst{2} \and
Jiaming Liu\inst{1} \\
Zhizhi Guo\inst{1} \and
Junyu Han\inst{1} \and
Jingtuo Liu\inst{1} \and
Errui Ding\inst{1} \and
Jingdong Wang\inst{1}}
\authorrunning{Z. Xu et al.}
%
\institute{Department of Computer Vision Technology (VIS), Baidu Inc. \and
S-Lab,  Nanyang Technological University\\
\email{\{xuzhiliang,zhouhang09,liujiaming03,dingerrui,wangjingdong\}@baidu.com} \\
\email{ziwei.liu@ntu.edu.sg}}
\maketitle


\input{sections/abstract}
\input{sections/intro}
\input{sections/related}
\input{sections/method}
\input{sections/experiments}
\input{sections/conclusion}

%
%
\bibliographystyle{splncs04}
\bibliography{egbib}
\clearpage
\appendix
\addcontentsline{toc}{section}{Appendices}

\section*{Appendices}

\section{Details of the Swapping-Driven ID Inversion}
\noindent\textbf{Strategy Revisit.}
Here we present more details of the \emph{Swapping-Driven ID Inversion} strategy. For clearer representation, we re-illustrate the pipeline in Fig.~\ref{fig:detail}. Inspired by the process of StyleGAN inversion, this strategy optimizes the features $\textbf{W}^+_{s} = \{\textbf{w}_{s(1)}, \dots, \textbf{w}_{s(2L)}\}$ in a total $N$ iterations.

At the $n$th iteration of the optimization, we denote the source identity-related $\mathcal{W}^+$ space feature as $\textbf{W}^{+\{n\}}_{s}$ and the desired output as $\textbf{W}^{+\{n + 1\}}_{s}$. $\textbf{W}^{+\{1\}}_{s}$ is initialized as $\{\textbf{w}_{s}, \dots, \textbf{w}_{s}\}$.
We randomly sample any image $I^{\{n\}}$, and firstly generate an intermediate frame $I^{r \rightarrow s}_g = \text{G}(\textbf{F}^{s}_{att}, f^r_{id})$. Then it is taken as the target frame to generate the cycled-back image 
\begin{align}
    I^{s \rightarrow r \rightarrow s}_g = \text{G}(\text{E}_{att}(I^{r \rightarrow s}_g), \textbf{W}^{+\{n\}}_{s})
\end{align}

The optimization is conducted using the identity loss $\mathcal{L}_{id}$ and the reconstruction loss $\mathcal{L}_{rec}$. We recap the two losses here. Given any source image $I_s$ and our generated image $I_g$, the identity loss between the two images are:
\begin{align}
    \mathcal{L}_{id}(I_g, I_s) = 1 - D_{\cos}(f^s_{id}, f^g_{id})),
\end{align}
where $D_{\cos}(f_a, f_b) = \frac{f_a\cdot f_b}{\|f_a\|_2\|f_b\|_2}$ denotes the cosine distance, $f^i_{id} = \text{E}_{id}(I_i)$ for any $I_i$. 
The reconstruction loss is:
\begin{align}
    \mathcal{L}_{rec}(I_g, I_s) &=\|{I}_{g} - {I}_{s} \|_1 +
    \sum_{m=1}^{N_{vgg}}\|\text{VGG}_m({I}_{g}) - \text{VGG}_{m}({I}_{s}) \|_1.
\end{align}

\noindent\textbf{Optimization Algorithm}
The choice of the final optimized $\mathcal{W}^+_s$ can be performed in two ways namely the \emph{one-to-one} optimization and \emph{one-to-many} optimization. The  \emph{one-to-one} optimization aims at finding the $\textbf{W}^+_{s}$ for swapping a specific source-target image pair ($I_s, I_t$), while \emph{one-to-one} optimization aims to find a general $\textbf{W}^+_{s}$ that is suitable for swapping the identity of $I_s$ to any face.

We start by introducing the  \emph{one-to-one} optimization. Within the total optimization iteration $N$, we select the $\textbf{W}^+_{s}$ as the feature that achieves the lowest $\mathcal{L}_{id}(I^{s \rightarrow t}_g, I_s)$. The whole optimization algorithm is depicted as follows:
\begin{algorithm}[h]
\caption{The algorithm of Swapping-Guided ID Inversion}
\label{algo:inversion}
\KwIn{A set of images with random identities $\{I^{\{1\}}_r, \dots, I^{\{N\}}_r\}$;   ~~~~~~~~~~~~~        The source image $I_s$; The trained encoders $\text{E}_{id}$, $\text{E}_{att}$ and G; ~~~~~~~~~~~The gradient-based optimizer $\mathcal{O}$.  The target image $I_t$.
}
\KwOut{The $\mathcal{W}^+$ space feature $\textbf{W}^+_s$.}
Initialize $\textbf{W}^{+}_s = \textbf{W}^{+\{1\}}_s = \{\textbf{w}_{s(1)}, \dots, \textbf{w}_{s(2L)}\}$, $\hat{\mathcal{L}}_{id} = 100$ and $n = 1$.

\While{$n \leq N$}{
   $I^{r \rightarrow s}_g \gets \text{G}(\text{E}_{att}(I_s), \text{E}_{id}(I^{\{n\}}_r))$\;
   $I^{s \rightarrow r \rightarrow s}_g \gets \text{G}(\text{E}_{att}(I^{r \rightarrow s}_g), \textbf{W}^{+\{n\}}_{s})$\;
   $\mathcal{L} \gets \lambda_{rec}\mathcal{L}_{rec}(I^{s \rightarrow r \rightarrow s}_g, I_s) + \lambda_{id}\mathcal{L}_{id}(I^{s \rightarrow r \rightarrow s}_g, I_s)$\;
   $I^{s \rightarrow t}_g \gets \text{G}(\text{E}_{att}(I_t), \textbf{W}^{+\{n\}}_s)$\;
    \If{$\mathcal{L}_{id}(I^{s \rightarrow t}_g, I_s) < \hat{\mathcal{L}}_{id}$}
    {
        $\textbf{W}^{+}_s \gets \textbf{W}^{+\{n\}}_s$\;
        $\hat{\mathcal{L}}_{id} \gets \mathcal{L}_{id}(I^{s \rightarrow t}_g)$ 
    }
   $\textbf{W}^{+\{n + 1\}}_s \gets \textbf{W}^{+\{n\}}_s - \eta * \mathcal{O}(\nabla_{_{ \textbf{W}}}\mathcal{L})$\;

   $n \gets n + 1$;
   }
\end{algorithm}

\setlength{\tabcolsep}{10pt}
\label{table:ff++}
\begin{table*}[t] 
\caption{\footnotesize \textbf{Face Forgery Detection Experiments.} The cross-dataset evaluation of enlarging the training set with FaceShifter's and our results}
\begin{tabular}{|l|c|c|c|c|c|}
\hline
Training set $\backslash$ Test set       & FF++	&Deepwild&	CDF &	DFDC&	Kodf\\
\hline
FF++ & 87.66 &65.89	&66.60 &67.62 &66.00 \\
FF++ w/ FaceShifter	&87.82 & 70.95 & 71.92 & 69.35 & 70.85 \\
\textbf{FF++ w/ Ours} & \textbf{88.16} & \textbf{70.96} & \textbf{73.59} & \textbf{70.07} & \textbf{72.13} \\
\hline
\end{tabular}
\end{table*}

$N$ is empirically selected as 200. After the optimization, the optimized $\textbf{W}^{+}_s$ can be sent into the generator for swapping any target face $I^{s \rightarrow t}_g = \text{G}(\text{E}_{att}(I_t), \textbf{W}^{+}_{s})$ given the source image $I_s$.

As for the \emph{one-to-many} optimization, all parts related to $\hat{\mathcal{L}}_{id}$ are not required. Thus $\textbf{W}^+_{s}$ is set as $\textbf{W}^{+\{N + 1\}}_{s}$. According to empirical studies carried out on the \emph{one-to-one} setting, the inversion procedure normally optimizes the identity similarity around the first $50$ iterations. Thus we set $N = 50$ in the \emph{one-to-many} setting, and this is the standard setting in our experiments.

\section{Experiments on Face Forgery Detection}
We conduct face forgery detection experiments with backbone Xception~\cite{chollet2017xception} that has been widely used as baseline in previous face forgery detection methods~\cite{roessler2019faceforensicspp,li2020face,qian2020thinking}. 
 The experiments are carried out on the following datasets.
\textbf{1)} FaceForensics++ (FF++)~\cite{roessler2019faceforensicspp} that has been introduced in the main paper. It is the most popular dataset used in face forgery detection.
\textbf{2)} WildDeepfake (Deepwild)~\cite{zi2020wilddeepfake} contains 3805 real clips and 3509 fake clips. All these videos are manually collected from the Internet. \textbf{3)} Celeb-DF (CDF)~\cite{li2020celeb} which contains high-quality face-swapped videos. \textbf{4)} DeepFake Detection Challenge (DFDC)~\cite{dolhansky2020deepfake} which is one of the most challenging datasets.

As for evaluation metrics, we use Area Under the Receiver Operating Characteristic Curve (AUC). The final confidence score of one video comes from the average of the first 80 frames. The baseline model is trained with 0/1 label (0 for real, 1 for fake, and p-fake) supervision using binary cross-entropy loss.

Specifically, our baseline model is trained on FF++~\cite{roessler2019faceforensicspp} without involving FaceShifter~\cite{li2019faceshifter} data. Then we additional involve 50,000 fake images generated by our method and 50,000 fake images from the results of FaceShifter to enlarge the training set to \textbf{FF++ w/ Ours}, and  \textbf{FF++ w/ FaceShifter} respectively. The results are shown in the Table~\ref{table:ff++}.

It can be seen that the model trained with the assistant of our method outperforms the model trained assisted with FaceShifter, which proves that our model could be more useful to the deepfake detection community. We suppose that it is because our model creates fewer artifacts and appears to be more realistic. Thus the forgery detection model trained on our data has more generalization ability.

\end{document}

%% file: sections/abstract.tex
\begin{abstract}
    Numerous attempts have been made to the task of person-agnostic face swapping given its wide applications. While existing methods mostly rely on tedious network and loss designs, they still struggle in the information balancing between the source and target faces, and tend to produce visible artifacts. In this work, we introduce a concise and effective framework named \textbf{StyleSwap}. Our core idea is to \emph{leverage a style-based generator to empower high-fidelity and robust face swapping, thus the generator's advantage can be adopted for optimizing identity similarity.}
    We identify that with only minimal modifications, a StyleGAN2 architecture can successfully handle the desired information from both source and target.
    Additionally, inspired by the ToRGB layers, a \emph{Swapping-Driven Mask Branch} is further devised to improve information blending.
    Furthermore, the advantage of StyleGAN inversion can be adopted. 
    Particularly, a \emph{Swapping-Guided ID Inversion} strategy is proposed to optimize identity similarity. Extensive experiments validate that our framework generates high-quality face swapping results that outperform state-of-the-art methods both qualitatively and quantitatively. Video, code, and models can be found at \url{https://hangz-nju-cuhk.github.io/projects/StyleSwap}.
\end{abstract}

\keywords{Face Swapping, Style-based Generator, GAN.}

%% file: sections/intro.tex
\begin{figure}
    \centering
    \includegraphics[width=1\textwidth]{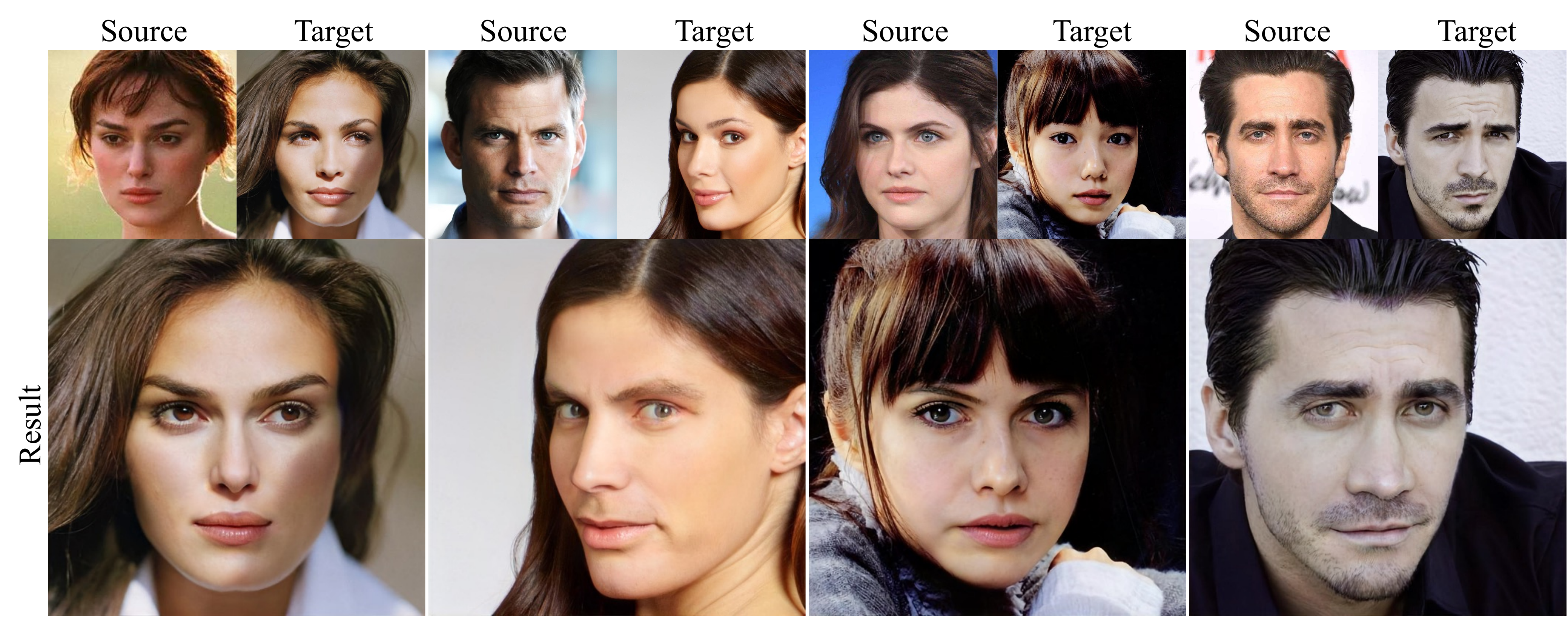}
    \caption{ \textbf{Qualitative results on $512 \times 512$ resolution.} Our method is robust under complicated conditions and achieves high-fidelity face swapping results}
        \vspace{-10pt}
    \label{fig:teaser}
\end{figure}

\section{Introduction}
The task of face swapping has drawn great attention~\cite{blanz2004exchanging,bitouk2008face,korshunova2017fast,perov2020deepfacelab,nguyen2019deep,nirkin2019fsgan,li2019faceshifter,zhu2020aot,wang2021hififace,xu2022mobilefaceswap} due to its wide applications in the fields of entertainment, film making, virtual human creation,  privacy protection, \emph{etc}. It aims at transferring the facial identity from a source person to a target frame or video while preserving attributes information including pose, expressions, lighting condition, and background~\cite{nguyen2019deep}. 

With the development of deep learning, generative models have been leveraged to boost face swapping quality~\cite{korshunova2017fast,natsume2018rsgan,nirkin2019fsgan,li2019faceshifter,wang2021hififace,xu2021facecontroller}.
However, the task is still challenging, particularly under the identity-agnostic setting where only a single frame is provided as the source image but targeting different sorts of scenarios.
The key challenges lie in two essential parts: how to explicitly capture the identity information; and how to blend the swapped face into the target seamlessly while preserving the implicit attributes unchanged. 

To tackle the above problems, previous studies take two different paths. 
\textbf{1)} Graphics-based methods~\cite{blanz2004exchanging,bitouk2008face} have involved the strong prior knowledge of intermediate structural representations such as landmarks and 3D models~\cite{egger20203d} into face swapping long ago.
%
Recent researchers combine this information with generative adversarial networks (GANs)~\cite{goodfellow2014generative} for identity and expression extraction~\cite{nirkin2018face,nirkin2019fsgan,xu2021facecontroller,wang2021hififace}. However, the inaccuracy of structural information would greatly influence the stableness and coherence of generated results, particularly in videos. \textbf{2)} Other methods explore pure learning-based pipelines~\cite{li2019faceshifter,chen2020simswap,gao2021information,Zhu_2021_CVPR}. Most of 
them rely on tedious loss and network structure designs~\cite{li2019faceshifter,chen2020simswap,gao2021information} for balancing the information between source and target images. 
Such designs make training difficult and fail in expressing desired information, which leads to non-similar or non-robust results with visible artifacts. 

Recently, StyleGAN architectures~\cite{karras2019style,karras2020analyzing,Karras2021} and their variants have been verified effective on various facial generative tasks,  including face attributes editing~\cite{abdal2019image2stylegan,abdal2020image2stylegan++,shen2020interfacegan,shen2021closedform}, face enhancement~\cite{wang2021gfpgan,yang2021gan}, and even face reenactment~\cite{burkov2020neural,zhou2021pose}. It is owing to style-based generator's strong expressibility and its advantages in latent space manipulation. But the exploration of such architectures in face swapping~\cite{Zhu_2021_CVPR,Xu_2022_CVPR} is still insufficient.
Specifically, 
the lighting conditions are greatly condemned in Zhu \textit{et al.}~\cite{Zhu_2021_CVPR} due to the limited distribution covered by the fixed generator. 
The structure of their feature blending procedure is also designed in a hand-crafted and layer-specific manner, which requires complicated human tuning. Concurrently, Xu \textit{et al.}~\cite{Xu_2022_CVPR}  aggregate the StyleGAN2 features with another designed encoder and decoder. Thus, a natural question arises: can we avoid tedious layer-by-layer structure design by adopting a versatile style-based generator~\cite{karras2019style,karras2020analyzing} with only minimal modifications? 

To this end, we propose \textbf{StyleSwap}, a concise and effective pipeline that empowers face swapping by a style-based generator. It produces results with higher fidelity, identity similarity and is more robust (\emph{i.e} avoids visible artifacts creation) under different scenarios compared with previous methods. Moreover, it is easy to implement and friendly for training.
The key is to \emph{adapt StyleGAN2 architecture to face swapping data flows through simple modifications, and adopt the generator's advantage for identity optimization.}
%
Detailedly, we first achieve the restoration of the target image's attributes with a simple layer-fusion strategy. The same idea has been proven to maintain the original StyleGAN's capability~\cite{yang2021gan}. Then we argue that the identity information can be injected by mapping extracted identity features to the $\mathcal{W}$ space. In this way, the identity information can be implicitly blended into the attributes in the convolution operations. 
Additionally, we propose a \emph{Swapping-Driven Mask Branch} which is identical to the \emph{ToRGB} branch. It naturally enforces the network to focus less on the target's high-level information and benefits final image blending. 


We further illustrate the advantages of this architecture by involving a simple optimization strategy for improving identity similarity. As the identity feature is mapped to the $\mathcal{W}$ space, a natural inspiration from recent StyleGAN inversion studies~\cite{abdal2019image2stylegan,abdal2020image2stylegan++} is to optimize a powerful $\mathcal{W^+}$ space through self-reconstruction. To avoid mode collapse, we introduce a novel \emph{Swapping-Guided ID Inversion} strategy by iteratively performing feature optimization and face swapping. 
Armed with these tools, we show that our StyleSwap generates high-fidelity results with simple video training paradigms. It is particularly robust 
and can be supported with enhanced data for generating high-resolution results. 

We summarize our contributions as follows: 
\textbf{(1)} We present the \textbf{StyleSwap} framework, which adopts a style-based generator into the person-agnostic face swapping task by simple modifications and the design of a \emph{Swapping-Driven Mask Branch}. It is easy to implement and train.
\textbf{(2)} By taking the advantage of the StyleGAN model, we design the novel \emph{Swapping-Guided ID Inversion} strategy to improve the identity similarity. 
\textbf{(3)} Extensive experiments demonstrate that our method outperforms the state of the arts on person-agnostic face swapping. Particularly, it demonstrates great robustness and has the capability of generating high-quality results.


%% file: sections/related.tex
\section{Related Work}
\subsection{Face Swapping}
\noindent\textbf{Structural Prior-Guided Face Swapping.} The task of face swapping has long been a research interest for both the computer graphics and computer vision community. Structural information such as 3D models and landmarks provide strong prior knowledge. 
Blanz \textit{et al.}~\cite{blanz2004exchanging} leverage 3DMM, and 
Bitouk \textit{et al.} use 3D lighting basis to design adjustment-based methods. Both of them rely on manual interaction, and can hardly change the source's expressions.  Nirkin \textit{et al.}~\cite{nirkin2018face} involve 3DMM with learned masks, but they render unrealistic results. 
Recent studies~\cite{nirkin2019fsgan,jiang2020deeperforensics,xu2021facecontroller,wang2021hififace} combine structural information with GANs for identity-agnostic face swapping. Xu \textit{et al.} and Wang \textit{et al.} both inject the parameters of 3DMM  into self-designed architectures.
Though high-fidelity results can be generated, the inaccuracy of 3D models and the need for inpainting greatly harm the temporal coherence and robustness of these methods under the video face swapping setting.

\noindent \textbf{Reconstruction-Based Face Swapping.} On the other hand, pure reconstruction based methods with GANs have also shown success. Korshunova \textit{et al.}~\cite{korshunova2017fast} train a network for swapping paired identities. The popular Deepfakes and DeepFaceLab~\cite{perov2020deepfacelab} share the same setting. However, these methods cannot generalize to arbitrary identities, which  limits their practical usage. 

As for person-agnostic face swapping, Li \textit{et al.}~\cite{li2019faceshifter} build the Faceshifter network. SimSwap~\cite{chen2020simswap} improves the expression consistency. However, they generate low-quality results with visible artifacts under certain circumstances. Recently, InfoSwap~\cite{gao2021information} creates high quality results by building a pipeline that relies on careful loss designs. It involves multi-stage of finetuning on various datasets. 
Different from these methods, we would like to ease the network design procedure with a style-based generator.

Specifically, Wang \textit{et al.}~\cite{wang2021one} firstly leverage a pretrained StyleGAN generator for high resolution face swapping. However, in order to adapt to the latent spaces of a pretrained StyleGAN generator, the authors design layer-specific fusion strategies, which involves a large number of hyper-parameters and ablative studies. Moreover, their method cannot keep the lighting conditions of the target frames. In our work, we retrain the style-based generator with simple modifications that preserve the attribute information better.

\subsection{ Facial Editing with Style-based Generator}

The strong ability of StyleGAN~\cite{karras2019style,karras2020analyzing,Karras2021} has been shown in various facial editing tasks including facial attributes editing~\cite{abdal2019image2stylegan,abdal2020image2stylegan++,shen2020interfacegan,shen2021closedform,tov2021designing,richardson2021encoding}, blind face restoration~\cite{wang2021gfpgan,yang2021gan}, face reenactment~\cite{burkov2020neural,zhou2021pose,Liang_2022_CVPR}, hairstyle editing~\cite{zhu2021barbershop}, and so on~\cite{zhu2021barbershop,tov2021designing,alaluf2021restyle,sun2021speech2talking}.

\noindent\textbf{StyleGAN Inversion.} Most face attribute editing framework~\cite{shen2020interfacegan,shen2021closedform} fix the pretrained generator unchanged and perform StyleGAN inversion. Abdal \textit{et al.}~\cite{abdal2019image2stylegan,abdal2020image2stylegan++} expand the original $\mathcal{W}$ latent space to the $\mathcal{W^+}$ space during inversion and achieve better image reconstruction results.
Recent studies invert images with StyleGAN specific encoders~\cite{tov2021designing,alaluf2021restyle,richardson2021encoding} for fast inversion. In our work, we take the inspiration of StyleGAN inversion and expand our identity feature to $\mathcal{W^+}$ space for boosting identity similarity. The usage of the StyleGAN specific encoders is left as a future work.

\noindent\textbf{Face Reenactment with Style-based Generator.} Face reenactment is very similar to face swapping and even serves as part of the face swapping procedure in certain methods~\cite{nirkin2019fsgan,perov2020deepfacelab,Shu_2022_CVPR}. The difference is that it aims at keeping the source image's identity and background. Burkov~\textit{et al.}~\cite{burkov2020neural} encode identity and expression information into the  $\mathcal{W}$ space and re-train the generator in a simple pipeline. Later studies~\cite{zhou2021pose,Liang_2022_CVPR} then expand this pipeline to the audio-driven setting. 


%% file: sections/method.tex
\begin{figure}[t]
    \centering
    \includegraphics[width=0.97\linewidth]{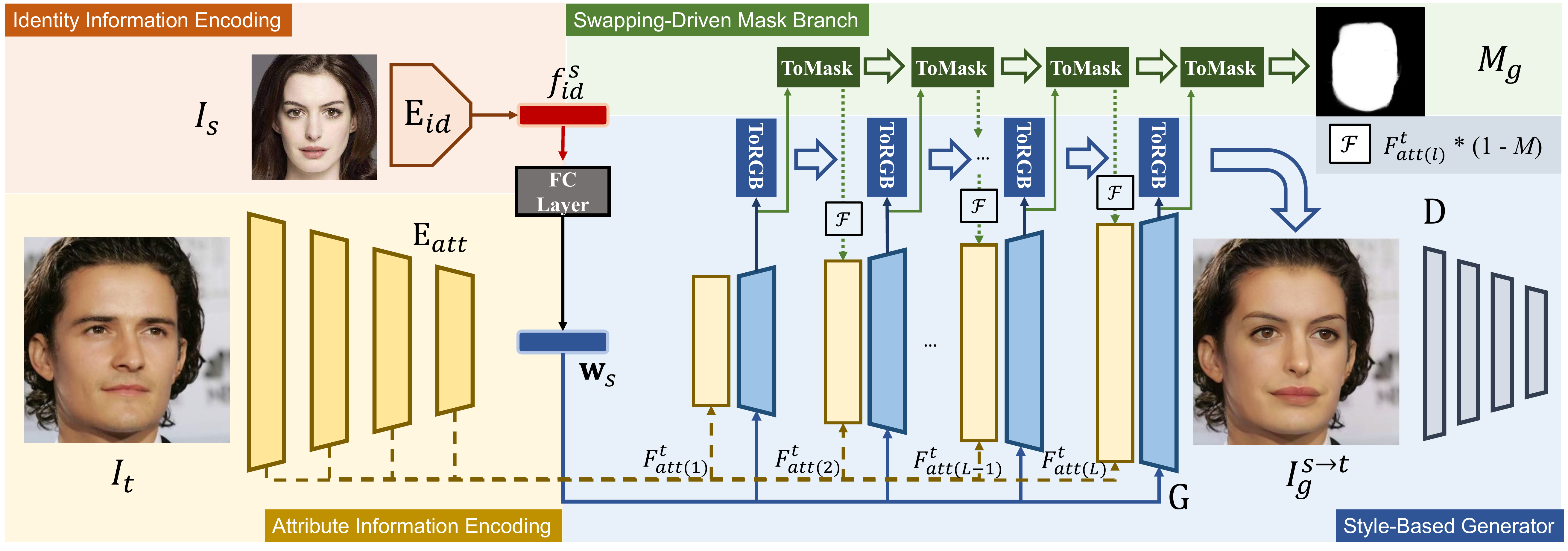}
    \caption{
   \textbf{Our StyleSwap framework.} The building blocks in \textbf{Blue} indicate the original structure of StyleGAN2. The source $I_s$ is encoded to $f^s_{id}$ by $\text{E}_{id}$ (\textbf{Red}). The attribute feature maps $\textbf{F}^t_{att}$ are encoded from $I_t$ by $\text{E}_{att}$ and concatenated to the StyleGAN2 generator blocks (\textbf{Yellow}). 
   Specifically, we devise a Swapping-Driven Mask Branch (\textbf{Green}) to predict a mask $M_g$ by leveraging  the same structure of ToRGB layers
}
\label{fig:pipeline}

\end{figure}

\section{Our Approach}

\noindent\textbf{Framework Overview.} The goal of person-agnostic face swapping is to 
swap the identity information from a source image $I_s$ onto a target frame $I_t$ while preserving $I_t$'s attribute information unchanged.

In this section, we present our \textbf{StyleSwap} framework (illustrated in Fig.~\ref{fig:pipeline}) which empowers face swapping by slightly modifying a style-based generator. 
In Section~\ref{sec:3.1} we introduce how to adapt a style-based generator to the face swapping task. Section~\ref{sec:3.2} illustrates the simple training paradigm. Importantly, in Section~\ref{sec:3.3}, we propose the \emph{Swapping-Guided ID Inversion} which specifically takes the advantage of GAN inversion for optimizing identity similarity.
%

\subsection{Adapting Style-Based Generator to Face Swapping}
\label{sec:3.1}

\noindent\textbf{Revisiting StyleGAN2.} We first recap the general setting of StyleGAN2~\cite{karras2020analyzing}. 
The original generator takes a constant 
feature map at the lowest resolution, then 
 a latent vector $\textbf{z}$ is sampled and mapped to a feature vector $\textbf{w}$ (lies in the $\mathcal{W}$ space). Afterwards, $\textbf{w}$ is sent into each layer of the $2L$-layer generator by affine transformations as the ``\emph{style}'' to modulate the convolutional kernels' weights. At each
resolution, a {ToRGB} layer is designed to render a three-channel RGB image progressively. The rough structures are depicted on Fig.~\ref{fig:pipeline} in \emph{blue}.

The attribute disentanglement ability in StyleGAN2 is implicitly achieved in the $\mathcal{W}$ or the expanded $\mathcal{W^+}$ space (if feeding different $\textbf{w}$ features to different layers).
As a result, operations on existing faces require inverting faces to latent vectors, which harms the preservation of spatial information. In our face swapping task, the problem lies in how to modify the generator so that the information of the target frame and identity can be sufficiently used.
%


\noindent\textbf{Infusing Attribute Information.} We propose to infuse the spatial information of the target frame as feature maps, rather than feature vectors, to preserve attribute information. 
A recent face restoration work~\cite{yang2021gan} verifies two important properties of StyleGAN2. Firstly, concatenating a noise map to each layer of the generator would not affect the network's generative ability. Secondly, such noise maps can be replaced by encoded spatial feature maps so that both the generative prior of StyleGAN2 model and the structural information of the input image can be kept.

Inspired by this observation, we apply a similar modification. Specifically, we leverage a simple encoder $\text{E}_{att}$ that encodes $I_t$ to different scales of spatial feature maps $\textbf{F}^t_{att} = \{F^t_{att(l)} | l \in [1, L]\}$. They are then infused into the StyleGAN2 architecture at each $2l$-th layer of the generator by concatenation following~\cite{yang2021gan}.

\noindent\textbf{Injecting Identity Feature.}  
In order to adapt to different target views, it is natural to encode the identity information into feature vectors. Here we use a pretrained identity encoder provided by ArcFace~\cite{deng2019arcface} to encode the identity feature $f^s_{id} = \text{E}_{id}(I_s)$. 

As the attributes on the face are already fused, we identify that modulated convolutions in StyleGAN2 are naturally suitable for the blending and shape-shifting of facial organs. 
%
Thus we directly map $f^s_{id}$ to $\textbf{w}_s$ in the $\mathcal{W}$ space ($\textbf{w}_s = \text{FC}_\textbf{w}(f^s_{id})$) through fully connected layers $\text{FC}_\textbf{w}$.

So far, a swapping result $I^{s \rightarrow t}_g = \text{G}(\textbf{F}^t_{att}, \textbf{w}_s)$ from the source to the target frame can already be rendered by the generator.

\noindent\textbf{Swapping-Driven Mask Branch.}
We then identify that learning a rough one-channel facial mask would benefit the whole face swapping step from two perspectives. 1) With a mask in the image domain, the areas that do not require modifications such as the background and hair can be directly kept unchanged. 2)  If the mask can be progressively and implicitly learned along with the generator, coarse masks at lower resolutions can help balance attribute and identity information in a similar way as~\cite{li2019faceshifter}.
We thus propose an additional modification by devising a \emph{Swapping-Driven Mask Branch} which takes the advantage of the StyleGAN2 model. Its structure is directly borrowed from the ``ToRGB'' branch, and denoted as ``ToMask''. Here we leverage a soft mask with values between $0$ and $1$.
The details of the mask branch are illustrated in Fig. 3 (a).

Let ${M_{g(l)}'}$ denote the one-channel output of the $l$-th ToMask network, which has the same resolution with the output of the $l$-th ToRGB layer and the $(l + 1)$-th $\textbf{F}^t_{att}$. The non-normalized mask $\tilde{M}_{g(l)}$ at the $l$-th layer of the mask branch is the combination of the ($l - 1$)-th layer's result and ${M_{g(l)}'}$. 
\begin{align}
\label{eq:1}
    \tilde{M}_{g(l)} = {upsample}(\tilde{M}_{g(l - 1)}) + {M_{g(l)}'},
\end{align}
where the bilinear $upsample$ is used. The softmasks we used are the normalized results ${M}_{g(l)} = \text{Sigmoid}(\tilde{M}_{g(l)})$. $M_{g(1)}$ is the normalized output of the first ToMask network, and $M_g$ is the final predicted mask. The face swapping result can be updated as:
\begin{align}
\label{eq:2}
    \hat{I}^{s \rightarrow t}_g = M_g * I^{s \rightarrow t}_g + (\textbf{1} - M_g) * I_t,
\end{align} 
where the $*$ denotes element-wise multiplication with broadcasting and  $\textbf{1}$ is the tensor with all ones. The mask repeats itself channel-wise three times to match the RGB channels.

\noindent\textbf{Masking Attribute Information.} 
\label{sec:3.1.2}
Our design above infuses facial attribute information into all layers. However, spatial information provided at  mid- and low-level layers might influence facial structures.
Though we expect the network to automatically perform information balancing, we identify that this procedure can be eased by blocking the attribute information relying on an implicitly learned mask. Thus we multiply our learned mask $M_{g(l)}$ at each resolution 
with the next layer's attribute feature map $F^t_{att(l + 1)}$ to an updated version: 
\begin{align}
\label{eq:3}
    \hat{F}^t_{att(l + 1)} = F^t_{att(l + 1)} * (\textbf{1} - M_{g(l)}).
\end{align}
Note that $M_{g(l)}$ and $F^t_{att(l + 1)}$ share the same spatial resolution, and there is no masking operation on $F^t_{att(1)}$ which provides the initial facial attributes. In this way, as $M_{g(l)}$ progressively grows to reach the ground truth mask, it also implicitly prevents the attribute information from influencing the identity similarity.  

\subsection{Training Paradigm}
\label{sec:3.2}

With our StyleSwap architecture, person-agnostic face swapping results can be learned in a simple end-to-end training paradigm. Particularly, we propose to involve certain video data for training a more robust swapping model. 

Given a source image $I_s$ and a target video $\{I_{T} | T \in \{t(1) \dots t(K)\}\}$, we generate the following frames: \textbf{(1)} The face swapping results from the source to any target frame: $I^{s \rightarrow t}_g = \text{G}(\textbf{F}^t_{att},\textbf{w}_s)$, where $\textbf{F}^t_{att} = \text{E}_{att}(I_t)$ and $\textbf{w}_s = \text{FC}_{\textbf{w}}(f_{id}^s) = \text{FC}_{\textbf{w}}(\text{E}_{id}(I_s))$. 
\textbf{(2)} The self-reconstruction results on the source frame itself: $I^{s \rightarrow s}_g = \text{G}(\textbf{F}^s_{att},\textbf{w}_s)$. \textbf{(3)} Particularly, we sample two target frames $I_{t(a)}$, $I_{t(b)}$ from a same video of a same person, and generate $I^{t(a) \rightarrow t(b)}_g = \text{G}(\textbf{F}^{t(b)}_{att}, \textbf{w}_{t(a)})$, $\textbf{w}_{t(a)} = \text{FC}_{\textbf{w}}(f^{t(a)}_{id})$.  

The training objectives consist of mainly four parts: the identity loss, feature matching loss, adversarial loss applied to all generated results, and the reconstruction loss applied to self- and cross-view reconstruction results. 

\noindent\textbf{Identity Loss.} The identity loss is built upon the cosine distances $D_{\cos}(f_a, f_b) = \frac{{f_a}^{\text{T}} \cdot f_b}{\|f_a\|_2\|f_b\|_2}$ between extracted identity features from $\text{E}_{id}$. Given any sampled data $I_i, I_j$ where $i,j \in \{s,T\}$:
\begin{align}
    \mathcal{L}_{id} = 1 - D_{\cos}(f^i_{id}, \text{E}_{id}(I^{i \rightarrow j}_g)).
\end{align}
Note that in order to disentangle identity information with illuminations, all images are augmented with color jittering when sent to the identity encoder.

\noindent\textbf{Adversarial Loss and Feature Matching Loss.} We directly adopt the original discriminator $\text{D}$ and adversarial loss functions of StyleGAN2~\cite{karras2019style}. For any $I^{i \rightarrow j}_g (i,j \in \{s,T\})$, $I_j$ is provided as the real image when applying the adversarial training. We denote this loss function as $\mathcal{L}_{adv}$ and omit the details.

Similar to~\cite{chen2020simswap}, we leverage a weak feature matching loss from the feature maps of the discriminator.
\begin{align}
    \mathcal{L}_{FM} = \sum_{m=n_{D}}^{N_{D}}(\|\text{D}_m(I^{i \rightarrow j}_g) -\text{D}_m(I_j) \|_1),
\end{align}
where $\text{D}_m$ denotes the $m$-th layer's output of the discriminator, and $n_{D}$ is the layer that starts computing the feature matching loss. This loss accounts for preserving the expression and certain low-level attribute information.

\noindent\textbf{Reconstruction Loss.} The reconstruction loss consists of the $L_1$ loss and the $\text{VGG}$ perceptual loss~\cite{wang2018pix2pixHD,park2019SPADE} when the pixel-level supervision can be provided. 
\begin{align}
    \mathcal{L}_{rec} &=\|{I}^{i  \rightarrow j}_{g} - {I}_{j} \|_1 +
    \sum_{m=1}^{N_{vgg}}\|\text{VGG}_m({I}^{i \rightarrow j}_{g}) - \text{VGG}_{m}({I}_{j}) \|_1,
\end{align}
where $\text{VGG}_{m}$ denotes the $m$-th layer's output of a  pre-trained VGG19 network. When applying the reconstruction loss, we set $i, j = {s}$ or $i , j \in \{T\}$. The reconstruction training provides supervision in the pixel space and has proven to be crucial in previous studies. Particularly, cross-view reconstruction is widely used in face reenactment training~\cite{kim2018deep,burkov2020neural,zhou2021pose}. It benefits  the attribute information's preservation by providing samples with different expressions on the source and target, and forces the network to learn with strong supervision.

\noindent\textbf{Mask Loss.} 
We leverage a pretrained facial mask predictor~\cite{sun2019high} to predict only a rough facial mask for each image in the training set. For each generated image $I^{i \rightarrow j}_{g} = \text{G}(\textbf{F}^j_{att},\textbf{w}_i)$, it is supervised with the binary mask of the target frame $M_j$.
The supervision can be written as:
\begin{align}
    \mathcal{L}_{mask} = \text{BCELoss}(M_g, M_j),
\end{align}
where BCELoss denotes the point-wise binary cross entropy loss. 

It is worth mentioning that the mask branch is a plug-in module that requires fine-tuning. When the \textbf{Masking Attribute Information} in Sec.~\ref{sec:3.1.2} is activated, all generated images are updated to $\hat{I}^{i \rightarrow j}_g$ as illustrated in Eq.~\ref{eq:2} and all $\textbf{F}_{att}$ are updated according to Eq.~\ref{eq:3}.

\noindent\textbf{Overall Loss Function.} The overall loss function for training the StyleSwap framework is the combination of the losses introduced above. It can be represented as:
\begin{align}
    \mathcal{L}_{total} =  \mathcal{L}_{adv} + \lambda_{id}\mathcal{L}_{id} + \lambda_{FM}\mathcal{L}_{FM} + \lambda_{rec}\mathcal{L}_{rec} + \lambda_{mask}\mathcal{L}_{mask},
\end{align}
where the $\lambda$s are balancing coefficients.

\begin{figure}[t]
    \centering
    \includegraphics[width=0.97\linewidth]{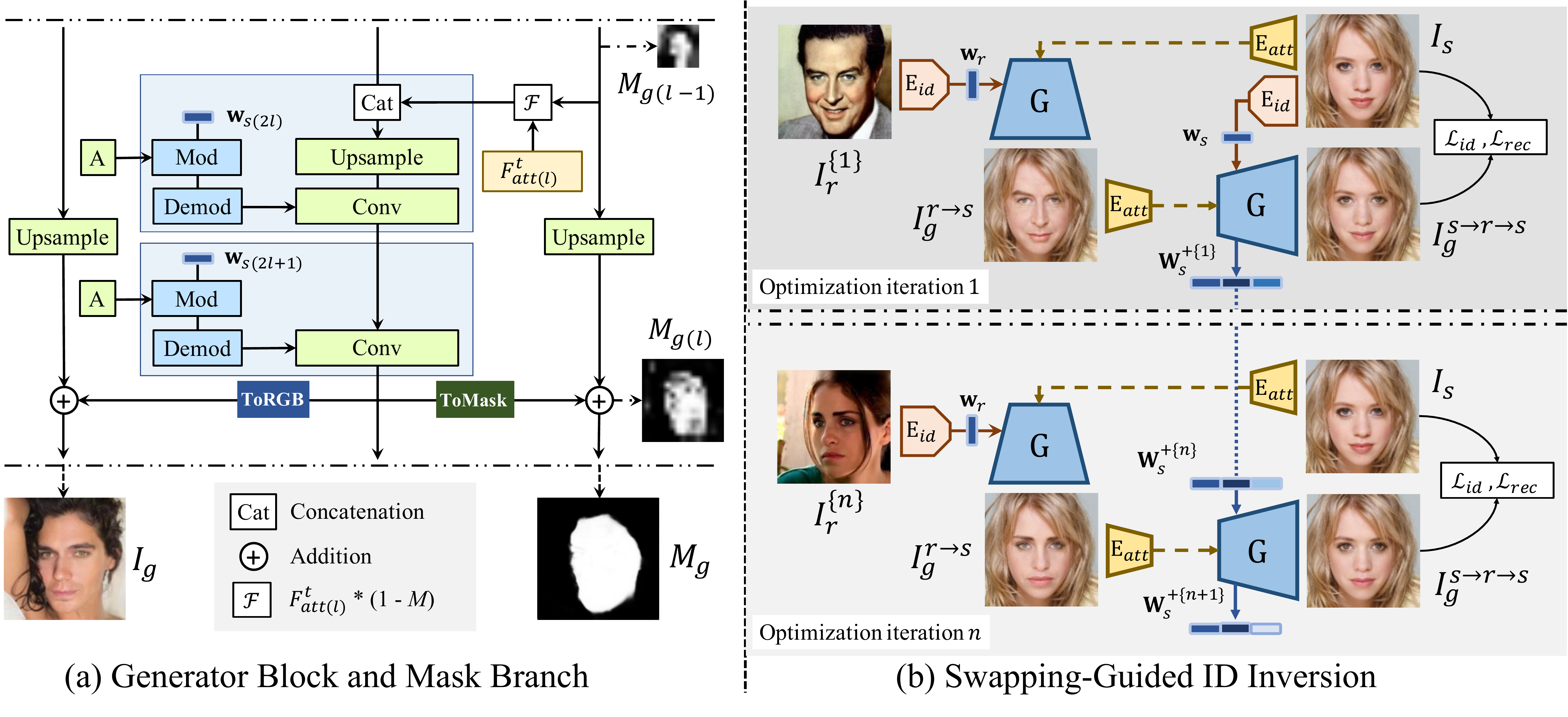}
    \caption{
  \textbf{(a)} The details of the generator blocks and the \emph{Swapping-Guided Mask Branch}. The figures show the block with visualized masks when $l = 4$ and the final outputs. $\textbf{w}_{s(2l)} = \textbf{w}_{s(2l + 1)} = \textbf{w}_{s}$ when the $\textbf{w}_{s}$ is not optimized. \textbf{(b)} The procedure for \emph{Swapping-Guided ID Inversion}. At each iteration, a different random $I^{\{n\}}_r$ is selected. The first and the $n$-th iterations are shown. $\textbf{w}_s$ is taken as initialization and updated to the $\mathcal{W}^+$ space
}
\label{fig:detail}
\end{figure}


\subsection{Swapping-Guided ID Inversion}
\label{sec:3.3}
Inspired by StyleGAN inversion, we illustrate one interesting property of our $\mathcal{W}$ space design: the identity similarity can be optimized in a GAN inversion manner through a \emph{Swapping-Guided ID Inversion} strategy, where the generator is fixed. 
As the encoded $\textbf{w}_s$ feature from the source lies in the $\mathcal{W}$ space, it can naturally be served as a good initialization.
We expect to optimize $\mathcal{W}$ or expanded $\mathcal{W}^+$ space for a specific person's identity.
The optimization procedure on $\mathcal{W^+}$ space is depicted in Fig.~\ref{fig:detail} (b).
The reconstruction loss $\mathcal{L}_{rec}$ and ID loss $\mathcal{L}_{id}$ are used.

\noindent\textbf{Challenges.} We first take $\mathcal{W}$ space optimization as an example, and expand it to the $\mathcal{W^+}$ space. An intuitive operation is to directly optimize $\textbf{w}_s$ (to any $\textbf{w}_s + \Delta\textbf{w}_s)$ using gradient descent at each self-reconstruction step $I^{s \rightarrow s}_g = \text{G}(\textbf{F}^{s}_{att},\textbf{w}_s + \Delta\textbf{w}_s)$. However, such practice would easily lead to mode collapse, \emph{i.e.}, direct reconstruction by ignoring the $\textbf{w}_s$ information. 
To tackle this problem, an alternative way is to perform face swapping before the optimization step (as shown in the optimization iteration $1$ in Fig.~\ref{fig:detail}). A randomly sampled face $I_r$ of arbitrary identity is used to build $I^{r \rightarrow s}_g$ which is later sent to the attribute encoder as the target frame. However, this way of restricting $I^{s \rightarrow r \rightarrow s}_g = \text{G}(I^{r \rightarrow s}_g, \textbf{w}_s)$ still relies on one fixed input image, which also leads to mode collapse during implementation.

\noindent\textbf{Iterative Identity Optimization and Swapping.} The key to our method is to feed a different face to the source image at each iteration of the optimization procedure. Specifically, at each iteration $n$, we randomly sample any $I^{\{n\}}_{r}$ to update the identity information on the source frame, so that the network perceives different identities. 

Moreover, a unique advantage of Style-based generator is to optimize the set of $2L$ different style features $\textbf{W}^+_s = [\textbf{w}_{s(1)}, \dots, \textbf{w}_{(2L)}]$ as performed in StyleGAN inversion studies~\cite{abdal2019image2stylegan,abdal2020image2stylegan++}. At each iteration, it is updated to $\textbf{W}^{+\{n\}}_s$.
After optimizing $\textbf{W}^{+}_s$, it can be directly leveraged by the generator to create face swapping result: $I^{s \rightarrow t}_g = \text{G}(\textbf{F}^t_{att}, \textbf{W}^{+}_s)$.
Such operation can hardly be performed on traditional face swapping pipelines, and has never been explored before. 
%
More details about the optimization algorithm are shown in the Supplementary Materials.

%% file: sections/experiments.tex
\section{Experiments}
\label{sec:4}
\noindent\textbf{Datasets.} We train our model on the VGGFace~\cite{parkhi2015deep} and a small part of VoxCeleb2~\cite{chung2018voxceleb2}. Due to the limitation of data quality, the original data from both datasets only supports  $256 \times 256$ resolution training. To show our capability of handling higher resolutions, we enhance the datasets with GPEN~\cite{yang2021gan} and train a $512 \times 512$ model. Part of the evaluation is conducted on FaceForensics++ (FF++) dataset~\cite{roessler2019faceforensicspp} which contains 1,000 Internet face videos and 1,000 Deepfakes~\cite{deepfakes} and 1000 official FaceShifter results. Evaluations on high-quality images leverage CelebA-HQ~\cite{karras2017progressive,liu2015deep}.  



\noindent\textbf{Implementation Details.} Our model is trained with batch size 64 on one NVIDIA Tesla A100 GPU for $256 \times 256$ resolution and batch size 12 for $512 \times 512$ resolution. We use the ADAM optimizer~\cite{kingma2014adam} with learning rate fixed at $1\times 10^{-4}$. The $\lambda_{id}$ is set at 10, $\lambda_{rec}$ and $\lambda_{FM}$ are set at 100. The other coefficients do not affect the generation results much thus empirically set at 1. 

A total of $2L = 14$ layers are used on $256 \times 256$ resolution images, and $2L = 16$ layers on $512 \times 512$ resolution. We leverage the modified ResNet50 provided by Arcface~\cite{deng2019arcface} as the identity encoder and fix it through out the experiments. 
The number of iterations required for optimizing the $\mathcal{W}$ and $\mathcal{W}^+$ for one identity is set empirically at $50$. 
Please refer to supplementary materials for more details.


\noindent\textbf{Competing Methods.} We compare our methods with previous state-of-the-arts reconstruction-based face swapping methods. They are \textbf{Faceshifter}~\cite{li2019faceshifter} and \textbf{SimSwap}~\cite{chen2020simswap} which leverage self-designed network structures; \textbf{InfoSwap}~\cite{gao2021information} that relies on tedious training procedure for high-resolution results; \textbf{MegaFS}~\cite{Zhu_2021_CVPR} which borrows a fixed StyleGAN2 pretrained model; and \textbf{Deepfakes}~\cite{deepfakes} which is a famous open-sourced tool. 
We use the official Deepfakes and FaceShifter results from the FF++ dataset and the officially released codes and models for producing swapped results of other methods.
We refer our results without identity optimization as \textbf{StyleSwap} and depict the optimized one separately as \textbf{StyleSwap w/ $ \mathcal{W}^+$}.

\setlength{\tabcolsep}{6pt}
\begin{table*}[t]
\caption{\textbf{Quantitative Results.} We report the percentage of successfully retrieved images on the \emph{ID Retrieval} metric. For the ID correlated metrics the higher the better, and it is the lower the better for other metrics}
\label{table:quantitative}
\begin{center}
\resizebox{\textwidth}{!}
{
\begin{tabular}{l|c| c| c| c|c}
\toprule
Method $\backslash$ Metric & ID Retrieval $\uparrow$& ID Cosine $\uparrow$  & Pose error$\downarrow$  & Exp. error $\downarrow$ & FID $\downarrow$ \\
\hline 
Deepfakes~\cite{deepfakes}  &  86.43\%  & 0.438 & 3.96 &  8.98 & 4.07 \\
FaceShiter~\cite{li2019faceshifter}  & 90.04\% & 0.510 & 2.19 & 6.77 &  3.50  \\
SimSwap~\cite{chen2020simswap}    & 93.07\%  & 0.578 & \textbf{1.36} &  \textbf{5.07} & 3.04 \\
MegaFS~\cite{Zhu_2021_CVPR}     & 89.12\% & 0.497 & 3.69 &  10.12 & 4.62\\
InfoSwap~\cite{gao2021information}     & 95.82\%   & 0.635 & 2.54 & 6.99  & 4.74 \\
\hline 
\textbf{StyleSwap (Ours)}     &  \textbf{97.05\%}  & \textbf{0.677} & 1.56 & 5.28 & \textbf{2.72} \\

\textbf{StyleSwap w/ $\mathcal{W}^+$}    &  \textbf{97.87\%}  & \textbf{0.706} & 1.51 & 5.27  & \textbf{2.58} \\
\bottomrule
\end{tabular}
}

\end{center}
\end{table*}


\begin{figure}[t]
    \centering
    \includegraphics[width=0.97\linewidth]{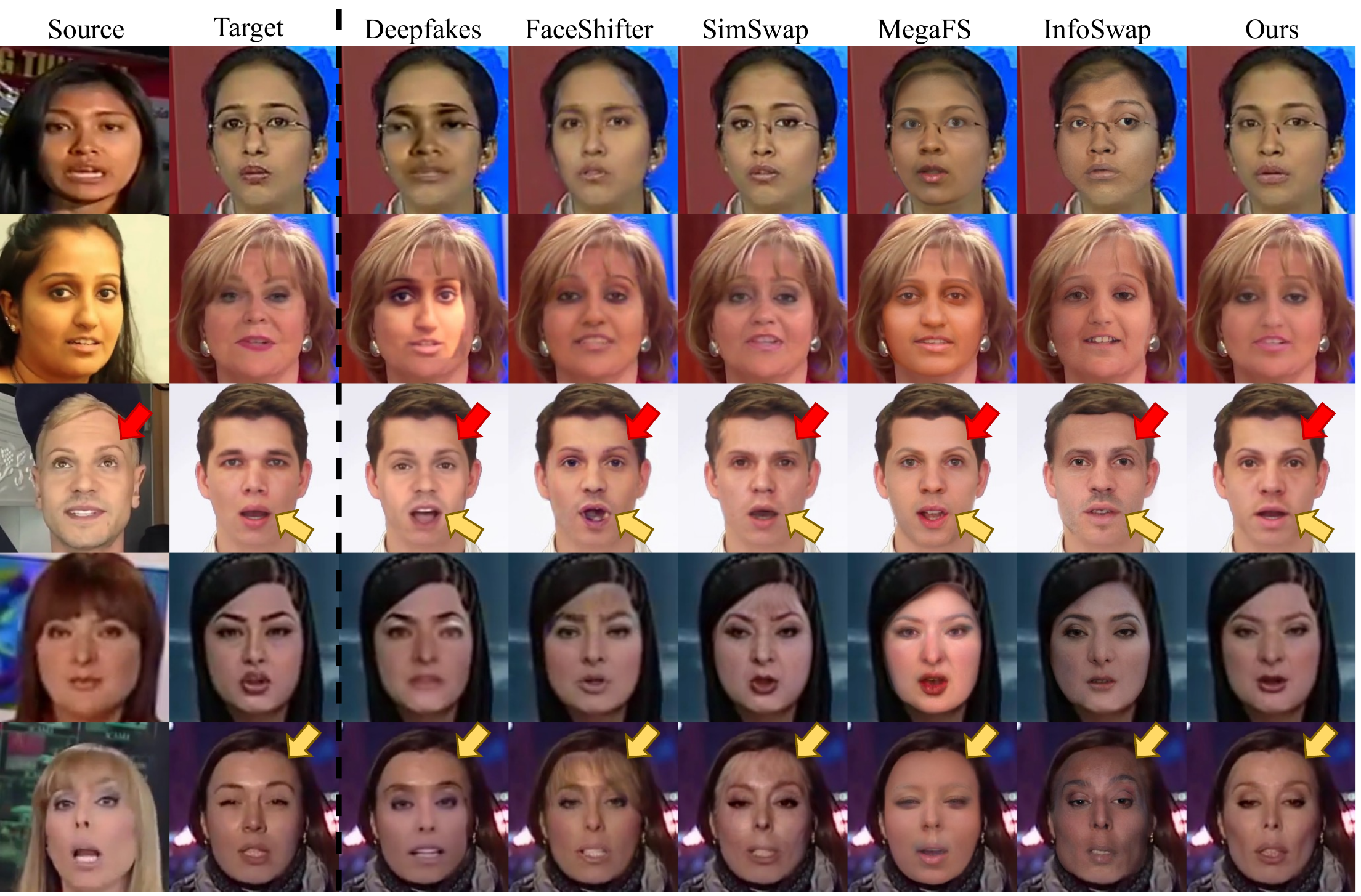}
    \caption{
  \textbf{Qualitative results on FF++ with $256 \times 256$ resolution.} Please pay attention to the identity information at the \textbf{red} arrows, and the attribute information at the \textbf{yellow} arrows
}
\label{fig:qualitative256}
\end{figure}


\subsection{Quantitative Evaluation}

\noindent\textbf{Evaluation Metrics.} The quantitative experiments are carried out on the FF++ dataset~\cite{roessler2019faceforensicspp}. Following~\cite{li2019faceshifter}, 10 frames are uniformly sampled
from the 1000 videos to get 10K faces for evaluation. 
We leverage five popularly used metrics.
The \emph{ID cosine similarity} and \emph{ID retrieval scores} are measured between swapped results and the source using another pretrained identity-recognition network~\cite{wang2018cosface}. Particularly, the calculation process of the ID retrieval score is the same as \cite{li2019faceshifter}.
The pose error is evaluated on the poses' $L_2$ distances produced by a pose estimator~\cite{ruiz2018fine}. The expression error is the $L_2$ distance between the results' and targets' expression embedding extracted from a facial expression extraction  model~\cite{vemulapalli2019compact}.

\noindent\textbf{Quantitative Results.} We list the results in Table~\ref{table:quantitative}. As can be seen that our \textbf{StyleSwap} model directly achieves state-of-the-art results on ID-related metrics and FID compared to previous ones, which shows the high similarity and image quality of our method. The \emph{Swapping-Driven ID Inversion} strategy further improves ID similarity. Meanwhile, we achieve comparable pose and expression errors with SimSwap~\cite{chen2020simswap} and outperform the others. The qualitative results further prove that we preserve the attribute information at a robust level.


\subsection{Qualitative Evaluation}

Subjective evaluation plays an important role in face swapping.  Note that for fair comparisons, we do not perform $\mathcal{W}^+$ space optimization. 
We provide image-based comparisons below. Demo videos and resources are available at \url{https://hangz-nju-cuhk.github.io/projects/StyleSwap}.

\noindent\textbf{Qualitative Results on $256 \times 256$ Resolution.} Here we show the image-based comparisons on $256 \times 256$ resolution in Fig.~\ref{fig:qualitative256}. The source and target figures are selected from FF++~\cite{roessler2019faceforensicspp}. It can be seen that MegaFS~\cite{Zhu_2021_CVPR} keeps the lighting conditions and texture badly, and InfoSwap~\cite{gao2021information} also fails to preserve accurate information. Though for some cases, SimSwap renders the best expression (the first row), it sometimes produces visible artifacts. Our method generates robust results with the highest similarity and competitive attribute preservation.

\begin{figure}[t]

    \centering
    \includegraphics[width=0.95\linewidth]{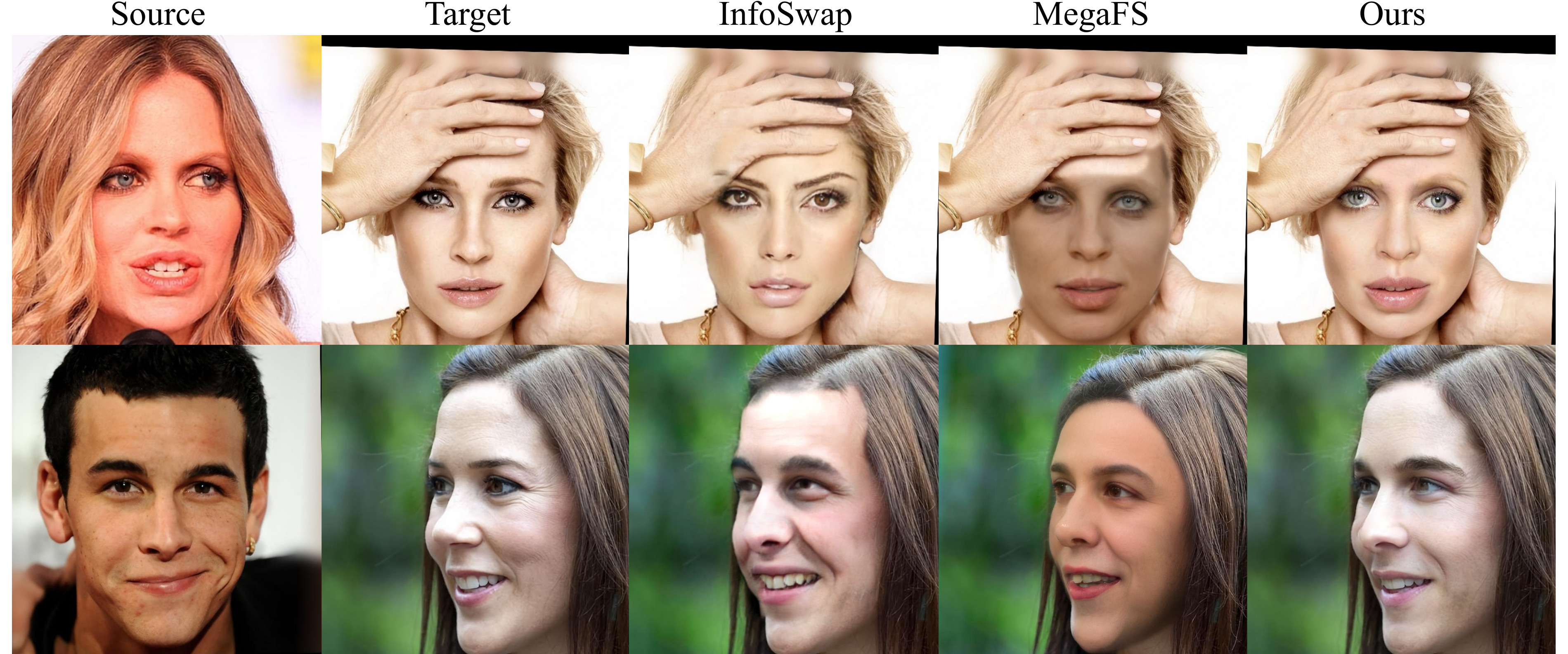}
    \caption{
  \textbf{Qualitative results} on CelebA-HQ with $512 \times 512$ resolution 
}
\label{fig:qualitative512}
\end{figure}

\noindent\textbf{Qualitative Results on $512 \times 512$ Resolution.} Particularly, we are able to generate high-resolution results with $512 \times 512$ resolution. Some results are shown in Fig.~\ref{fig:teaser}.
We compare our results with MegaFS~\cite{Zhu_2021_CVPR} and InfoSwap~\cite{gao2021information} which are also able to produce high-resolution results. 
As shown in Fig.~\ref{fig:qualitative512}, MegaFS generates a false skin tune of the target person and InfoSwap generates visible artifacts. Our method clearly outperforms theirs on high-resolution results.

\setlength{\tabcolsep}{2.2pt} 
\begin{table*}[t]  \small 
\begin{center}

\caption{\textbf{User Study's Ranking Scores.} Larger is higher, with the maximum value to be 5}
\label{table:usr}
\begin{tabular}{cccccc}

\hline

Perspective$\setminus$ Method & Faceshiter~\cite{li2019faceshifter} & SimSwap~\cite{chen2020simswap}  & MegaFS~\cite{Zhu_2021_CVPR} & InfoSwap~\cite{gao2021information} & \textbf{Ours} \\
\noalign{\smallskip}
\hline

ID Similarity & 3.05 & 2.59 &2.80& 2.89 & \textbf{3.68}\\
Att. Preservation &3.16 & 2.94 &2.21 &3.36 & \textbf{3.82}\\

Naturalness& 2.95& 2.62 & 2.43 &3.11 &\textbf{3.88}\\
\hline
\end{tabular}
\end{center}
\end{table*}
\setlength{\tabcolsep}{1.4pt}


\noindent\textbf{User Study.} We further conduct a user study for subjective evaluations. A total of 30 users are involved to discriminate 20 different samples from the FF++ dataset. The users are asked to rank the quality of the fake images from the following perspectives: \textbf{1) Id similarity} with the source image; \textbf{2) Attribute (Att) Preservation} including expressions and backgrounds referring to the target; and \textbf{3) Naturalness.} Whether there are visible artifacts on the face and does the figure look like a real person. Detailed instructions and training are provided. As Deepfakes cannot produce plausible results, we only conduct the user study on the other 4 methods and our StyleSwap. Thus we define the highest score to be 5 and the lowest score to be 1 for each case.

The results are shown in Table~\ref{table:usr}. All ranks are given corresponding scores with 5 being the highest. It can be seen that our method performs the best on given all metrics. Specifically, the low score of SimSwap with respect to attribute preservation is led by their artifacts.

\setlength{\tabcolsep}{8pt}
\begin{table*}[t] 
\caption{\textbf{Quantitative Ablation Study.} The StyleSwap model with $\mathcal{W}^+$ space optimization achieves the best results on all metrics}
\label{table:ablation}
\begin{center}
\resizebox{\textwidth}{!}
{
\begin{tabular}{l|c| c| c| c|c}
\toprule
Method $\backslash$ Metric & ID Retrieval $\uparrow$& ID Cosine $\uparrow$  & Pose error$\downarrow$  & Exp. error $\downarrow$ & FID $\downarrow$ \\
\hline 
StyleSwap (vector)    & 96.68\%  & 0.668 & 2.81 &  8.44 & 4.54 \\
StyleSwap w/o Mask     & 95.56\% & 0.653 & 1.80 &  5.78 & 3.18\\

StyleSwap     &  97.05\%  & 0.677 & 1.56 & 5.28 & 2.72 \\
{StyleSwap} w/ $\mathcal{W}$     &  97.54\%  & 0.693 & 1.52 & 5.28 & 2.60 \\

\textbf{StyleSwap w/ $\mathcal{W}^+$}    &  \textbf{97.87\%}  & \textbf{0.706} & \textbf{1.51} & \textbf{5.27}  & \textbf{2.58} \\
\bottomrule
\end{tabular}
}

\end{center}
\end{table*}


\begin{figure}[t]
    \centering
    \includegraphics[width=0.96\linewidth]{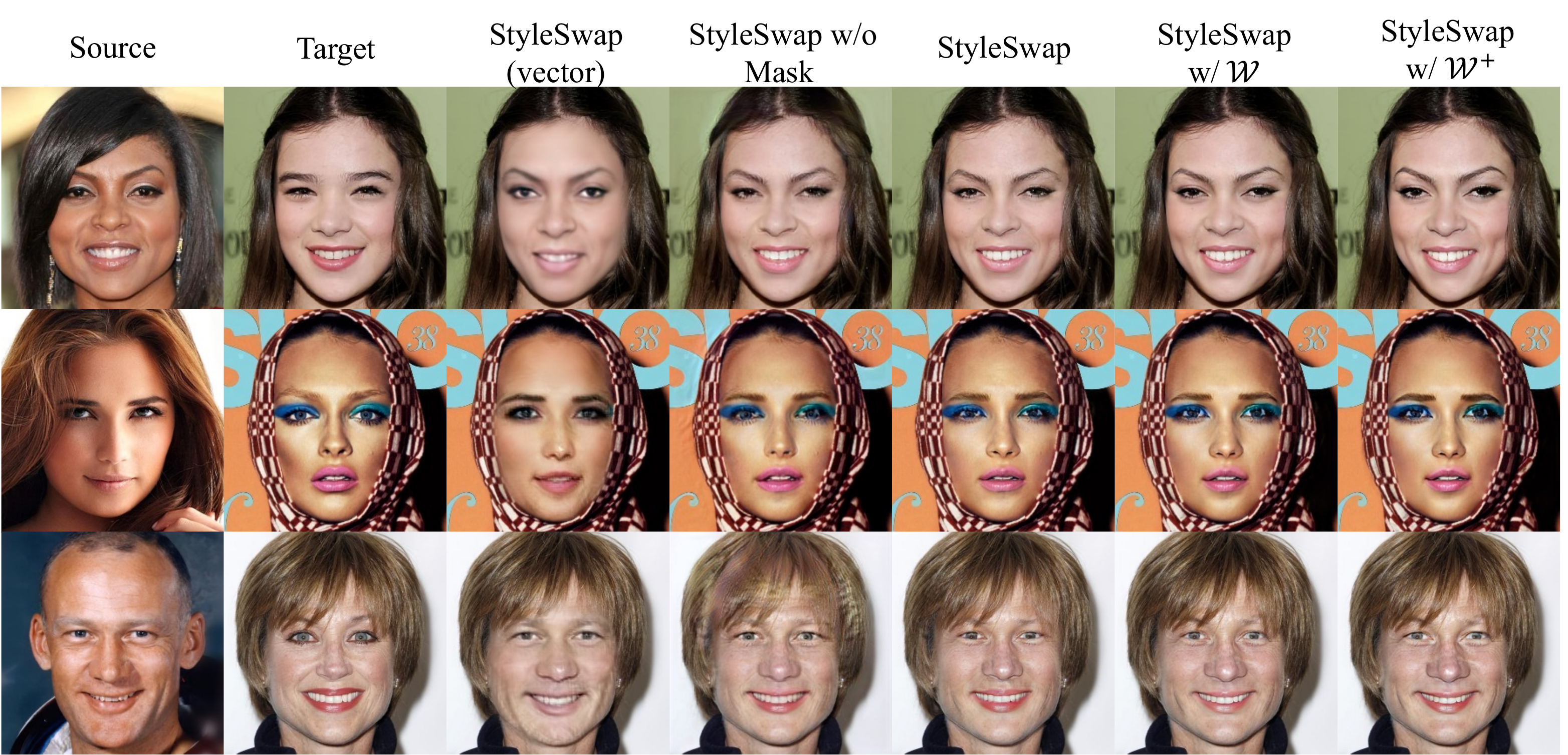}
    \caption{
  \textbf{ Qualitative Ablation Study on CelebA-HQ.
}}
\label{fig:ablation}
\end{figure}


\subsection{Further Analysis}
\noindent\textbf{Ablation Studies.} We conduct ablation studies on several important designs of our network. (1) The effectiveness of the $\mathcal{W}^+$ space optimization and its comparison with the $\mathcal{W}$ space optimization \textbf{StyleSwap w/ $\mathcal{W}$}; (2) Our StyleSwap without the {Mask Branch} \textbf{StyleSwap w/o Mask}. (3) Our design of concatenating the target's spatial feature maps into the style-based generator. An alternative way is to map both source and target information into the $\mathcal{W}$ as performed in~\cite{burkov2020neural,zhou2021pose}. This model is denoted as \textbf{StyleSwap (vector)}. The cross-reconstruction training paradigm does not affect numerical results and visualization in most cases, thus the comparison will be given. 

We show the quantitative ablation results in Table~\ref{table:ablation} under the same evaluation protocols on FF++, and the qualitative ablation results on Fig.~\ref{fig:ablation} which is generated on CelebA-HQ~\cite{karras2017progressive}. It can be seen that without the spatial feature maps, StyleSwap (vector) cannot preserve the attribute information correctly and thus loses details. Moreover, the identity similarity improves along with the addition of the \emph{Mask Branch}, the $\mathcal{W}$ and $\mathcal{W}^+$  space optimization.

\noindent\textbf{Effectiveness on Face Forgery Detection.} Importantly, our method can contribute to the face forgery detection community~\cite{roessler2019faceforensicspp,nguyen2019deep,guan2022delving}. We conduct experiments under the standard face forgery detection pipeline and additionally provide the same amount of generated data by our method and FaceShifter~\cite{li2019faceshifter}. The evaluation results on 4 different datasets validate that our method assists the forgery detection better. The details can be found in the Supplementary Materials.



%% file: sections/conclusion.tex
\section{Conclusion and Discussion}

\noindent\textbf{Conclusion.} In this paper, we propose \textbf{StyleSwap}, a concise and effective framework that adapts a style-based generator for high-fidelity face swapping. We emphasize several key properties of our method: 1) With only minor modifications to the StyleGAN2 generator, our method is easy to implement and friendly to train, which saves a lot of human labor. 2) With the strong capability of the style-based generator and the simple design of the \emph{Swapping-Guided Mask Branch}, our results are not only with high quality, similarity but enjoy high robustness. 3) Our method can take the advantage of GAN inversion and optimize the $\mathcal{W}^+$ space for improving identity similarity. 4) Our method can benefit face forgery detection by providing realistic fake results.

\noindent\textbf{Broader Impact.} Face swapping technique could create deepfake results for malicious purposes. We also take this issue into serious consideration and show the effectiveness of our method in the face forgery detection community. We will strictly limit the usage of this work for research purposes only.

\noindent\textbf{Acknowledgements.} This work is supported by NTU NAP, MOE AcRF Tier 1 (2021-T1-001-088), and under the RIE2020 Industry Alignment Fund – Industry Collaboration Projects (IAF-ICP) Funding Initiative, as well as cash and in-kind contributions from the industry partner(s).

%% file: main.bbl
\begin{thebibliography}{10}
\providecommand{\url}[1]{\texttt{#1}}
\providecommand{\urlprefix}{URL }
\providecommand{\doi}[1]{https://doi.org/#1}

\bibitem{abdal2019image2stylegan}
Abdal, R., Qin, Y., Wonka, P.: Image2stylegan: How to embed images into the
  stylegan latent space? In: Proceedings of the IEEE/CVF International
  Conference on Computer Vision. pp. 4432--4441 (2019)

\bibitem{abdal2020image2stylegan++}
Abdal, R., Qin, Y., Wonka, P.: Image2stylegan++: How to edit the embedded
  images? In: Proceedings of the IEEE/CVF Conference on Computer Vision and
  Pattern Recognition. pp. 8296--8305 (2020)

\bibitem{alaluf2021restyle}
Alaluf, Y., Patashnik, O., Cohen-Or, D.: Restyle: A residual-based stylegan
  encoder via iterative refinement. In: Proceedings of the IEEE/CVF
  International Conference on Computer Vision. pp. 6711--6720 (2021)

\bibitem{bitouk2008face}
Bitouk, D., Kumar, N., Dhillon, S., Belhumeur, P., Nayar, S.K.: Face swapping:
  automatically replacing faces in photographs. In: ACM SIGGRAPH 2008 papers,
  pp.~1--8 (2008)

\bibitem{blanz2004exchanging}
Blanz, V., Scherbaum, K., Vetter, T., Seidel, H.P.: Exchanging faces in images.
  In: Computer Graphics Forum. vol.~23, pp. 669--676. Wiley Online Library
  (2004)

\bibitem{burkov2020neural}
Burkov, E., Pasechnik, I., Grigorev, A., Lempitsky, V.: Neural head reenactment
  with latent pose descriptors. In: Proceedings of the IEEE/CVF conference on
  computer vision and pattern recognition. pp. 13786--13795 (2020)

\bibitem{chen2020simswap}
Chen, R., Chen, X., Ni, B., Ge, Y.: Simswap: An efficient framework for high
  fidelity face swapping. In: Proceedings of the 28th ACM International
  Conference on Multimedia. pp. 2003--2011 (2020)

\bibitem{chollet2017xception}
Chollet, F.: Xception: Deep learning with depthwise separable convolutions. In:
  Proceedings of the IEEE conference on computer vision and pattern
  recognition. pp. 1251--1258 (2017)

\bibitem{chung2018voxceleb2}
Chung, J.S., Nagrani, A., Zisserman, A.: Voxceleb2: Deep speaker recognition.
  arXiv preprint arXiv:1806.05622  (2018)

\bibitem{deepfakes}
Deepfakes: Faceswap. \url{https://github.com/deepfakes/faceswap}

\bibitem{deng2019arcface}
Deng, J., Guo, J., Xue, N., Zafeiriou, S.: Arcface: Additive angular margin
  loss for deep face recognition. In: Proceedings of the IEEE/CVF conference on
  computer vision and pattern recognition. pp. 4690--4699 (2019)

\bibitem{dolhansky2020deepfake}
Dolhansky, B., Bitton, J., Pflaum, B., Lu, J., Howes, R., Wang, M., Ferrer,
  C.C.: The deepfake detection challenge (dfdc) dataset. arXiv preprint
  arXiv:2006.07397  (2020)

\bibitem{egger20203d}
Egger, B., Smith, W.A., Tewari, A., Wuhrer, S., Zollhoefer, M., Beeler, T.,
  Bernard, F., Bolkart, T., Kortylewski, A., Romdhani, S., et~al.: 3d morphable
  face models—past, present, and future. ACM Transactions on Graphics (TOG)
  \textbf{39}(5),  1--38 (2020)

\bibitem{gao2021information}
Gao, G., Huang, H., Fu, C., Li, Z., He, R.: Information bottleneck
  disentanglement for identity swapping. In: Proceedings of the IEEE/CVF
  Conference on Computer Vision and Pattern Recognition. pp. 3404--3413 (2021)

\bibitem{goodfellow2014generative}
Goodfellow, I., Pouget-Abadie, J., Mirza, M., Xu, B., Warde-Farley, D., Ozair,
  S., Courville, A., Bengio, Y.: Generative adversarial nets. Advances in
  neural information processing systems  \textbf{27} (2014)

\bibitem{guan2022delving}
Guan, J., Zhou, H., Hong, Z., Ding, E., Wang, J., Quan, C., Zhao, Y.: Delving
  into sequential patches for deepfake detection. Advances in Neural
  Information Processing Systems  (2022)

\bibitem{jiang2020deeperforensics}
Jiang, L., Li, R., Wu, W., Qian, C., Loy, C.C.: Deeperforensics-1.0: A
  large-scale dataset for real-world face forgery detection. In: Proceedings of
  the IEEE/CVF conference on computer vision and pattern recognition. pp.
  2889--2898 (2020)

\bibitem{karras2017progressive}
Karras, T., Aila, T., Laine, S., Lehtinen, J.: Progressive growing of gans for
  improved quality, stability, and variation. arXiv preprint arXiv:1710.10196
  (2017)

\bibitem{Karras2021}
Karras, T., Aittala, M., Laine, S., H\"ark\"onen, E., Hellsten, J., Lehtinen,
  J., Aila, T.: Alias-free generative adversarial networks. In: Proc. NeurIPS
  (2021)

\bibitem{karras2019style}
Karras, T., Laine, S., Aila, T.: A style-based generator architecture for
  generative adversarial networks. In: Proceedings of the IEEE/CVF conference
  on computer vision and pattern recognition. pp. 4401--4410 (2019)

\bibitem{karras2020analyzing}
Karras, T., Laine, S., Aittala, M., Hellsten, J., Lehtinen, J., Aila, T.:
  Analyzing and improving the image quality of stylegan. In: Proceedings of the
  IEEE/CVF conference on computer vision and pattern recognition. pp.
  8110--8119 (2020)

\bibitem{kim2018deep}
Kim, H., Garrido, P., Tewari, A., Xu, W., Thies, J., Niessner, M., P{\'e}rez,
  P., Richardt, C., Zollh{\"o}fer, M., Theobalt, C.: Deep video portraits. ACM
  Transactions on Graphics (TOG)  (2018)

\bibitem{kingma2014adam}
Kingma, D.P., Ba, J.: Adam: A method for stochastic optimization. arXiv
  preprint arXiv:1412.6980  (2014)

\bibitem{korshunova2017fast}
Korshunova, I., Shi, W., Dambre, J., Theis, L.: Fast face-swap using
  convolutional neural networks. In: Proceedings of the IEEE international
  conference on computer vision. pp. 3677--3685 (2017)

\bibitem{li2019faceshifter}
Li, L., Bao, J., Yang, H., Chen, D., Wen, F.: Faceshifter: Towards high
  fidelity and occlusion aware face swapping. CVPR  (2020)

\bibitem{li2020face}
Li, L., Bao, J., Zhang, T., Yang, H., Chen, D., Wen, F., Guo, B.: Face x-ray
  for more general face forgery detection. In: Proceedings of the IEEE/CVF
  conference on computer vision and pattern recognition. pp. 5001--5010 (2020)

\bibitem{li2020celeb}
Li, Y., Yang, X., Sun, P., Qi, H., Lyu, S.: Celeb-df: A large-scale challenging
  dataset for deepfake forensics. In: Proceedings of the IEEE/CVF Conference on
  Computer Vision and Pattern Recognition. pp. 3207--3216 (2020)

\bibitem{Liang_2022_CVPR}
Liang, B., Pan, Y., Guo, Z., Zhou, H., Hong, Z., Han, X., Han, J., Liu, J.,
  Ding, E., Wang, J.: Expressive talking head generation with granular
  audio-visual control. In: Proceedings of the IEEE/CVF Conference on Computer
  Vision and Pattern Recognition (CVPR). pp. 3387--3396 (June 2022)

\bibitem{liu2015deep}
Liu, Z., Luo, P., Wang, X., Tang, X.: Deep learning face attributes in the
  wild. In: Proceedings of the IEEE international conference on computer
  vision. pp. 3730--3738 (2015)

\bibitem{natsume2018rsgan}
Natsume, R., Yatagawa, T., Morishima, S.: Rsgan: face swapping and editing
  using face and hair representation in latent spaces. arXiv preprint
  arXiv:1804.03447  (2018)

\bibitem{nguyen2019deep}
Nguyen, T.T., Nguyen, Q.V.H., Nguyen, C.M., Nguyen, D., Nguyen, D.T.,
  Nahavandi, S.: Deep learning for deepfakes creation and detection: A survey.
  arXiv preprint arXiv:1909.11573  (2019)

\bibitem{nirkin2019fsgan}
Nirkin, Y., Keller, Y., Hassner, T.: Fsgan: Subject agnostic face swapping and
  reenactment. In: Proceedings of the IEEE/CVF international conference on
  computer vision. pp. 7184--7193 (2019)

\bibitem{nirkin2018face}
Nirkin, Y., Masi, I., Tuan, A.T., Hassner, T., Medioni, G.: On face
  segmentation, face swapping, and face perception. In: 2018 13th IEEE
  International Conference on Automatic Face \& Gesture Recognition (FG 2018).
  pp. 98--105. IEEE (2018)

\bibitem{park2019SPADE}
Park, T., Liu, M.Y., Wang, T.C., Zhu, J.Y.: Semantic image synthesis with
  spatially-adaptive normalization. In: Proceedings of the IEEE Conference on
  Computer Vision and Pattern Recognition (2019)

\bibitem{parkhi2015deep}
Parkhi, O.M., Vedaldi, A., Zisserman, A.: Deep face recognition  (2015)

\bibitem{perov2020deepfacelab}
Perov, I., Gao, D., Chervoniy, N., Liu, K., Marangonda, S., Um{\'e}, C., Dpfks,
  M., Facenheim, C.S., RP, L., Jiang, J., et~al.: Deepfacelab: Integrated,
  flexible and extensible face-swapping framework. arXiv preprint
  arXiv:2005.05535  (2020)

\bibitem{qian2020thinking}
Qian, Y., Yin, G., Sheng, L., Chen, Z., Shao, J.: Thinking in frequency: Face
  forgery detection by mining frequency-aware clues. In: European Conference on
  Computer Vision. pp. 86--103. Springer (2020)

\bibitem{richardson2021encoding}
Richardson, E., Alaluf, Y., Patashnik, O., Nitzan, Y., Azar, Y., Shapiro, S.,
  Cohen-Or, D.: Encoding in style: a stylegan encoder for image-to-image
  translation. In: Proceedings of the IEEE/CVF Conference on Computer Vision
  and Pattern Recognition. pp. 2287--2296 (2021)

\bibitem{roessler2019faceforensicspp}
R\"ossler, A., Cozzolino, D., Verdoliva, L., Riess, C., Thies, J., Nie{\ss}ner,
  M.: Face{F}orensics++: Learning to detect manipulated facial images. In:
  International Conference on Computer Vision (ICCV) (2019)

\bibitem{ruiz2018fine}
Ruiz, N., Chong, E., Rehg, J.M.: Fine-grained head pose estimation without
  keypoints. In: Proceedings of the IEEE conference on computer vision and
  pattern recognition workshops. pp. 2074--2083 (2018)

\bibitem{shen2020interfacegan}
Shen, Y., Yang, C., Tang, X., Zhou, B.: Interfacegan: Interpreting the
  disentangled face representation learned by gans. TPAMI  (2020)

\bibitem{shen2021closedform}
Shen, Y., Zhou, B.: Closed-form factorization of latent semantics in gans. In:
  CVPR (2021)

\bibitem{Shu_2022_CVPR}
Shu, C., Wu, H., Zhou, H., Liu, J., Hong, Z., Ding, C., Han, J., Liu, J., Ding,
  E., Wang, J.: Few-shot head swapping in the wild. In: Proceedings of the
  IEEE/CVF Conference on Computer Vision and Pattern Recognition (CVPR). pp.
  10789--10798 (June 2022)

\bibitem{sun2019high}
Sun, K., Zhao, Y., Jiang, B., Cheng, T., Xiao, B., Liu, D., Mu, Y., Wang, X.,
  Liu, W., Wang, J.: High-resolution representations for labeling pixels and
  regions. arXiv preprint arXiv:1904.04514  (2019)

\bibitem{sun2021speech2talking}
Sun, Y., Zhou, H., Liu, Z., Koike, H.: Speech2talking-face: Inferring and
  driving a face with synchronized audio-visual representation. In: IJCAI.
  vol.~2, p.~4 (2021)

\bibitem{tov2021designing}
Tov, O., Alaluf, Y., Nitzan, Y., Patashnik, O., Cohen-Or, D.: Designing an
  encoder for stylegan image manipulation. ACM Transactions on Graphics (TOG)
  (2021)

\bibitem{vemulapalli2019compact}
Vemulapalli, R., Agarwala, A.: A compact embedding for facial expression
  similarity. In: Proceedings of the IEEE/CVF Conference on Computer Vision and
  Pattern Recognition. pp. 5683--5692 (2019)

\bibitem{wang2018cosface}
Wang, H., Wang, Y., Zhou, Z., Ji, X., Gong, D., Zhou, J., Li, Z., Liu, W.:
  Cosface: Large margin cosine loss for deep face recognition. In: Proceedings
  of the IEEE conference on computer vision and pattern recognition. pp.
  5265--5274 (2018)

\bibitem{wang2018pix2pixHD}
Wang, T.C., Liu, M.Y., Zhu, J.Y., Tao, A., Kautz, J., Catanzaro, B.:
  High-resolution image synthesis and semantic manipulation with conditional
  gans. In: Proceedings of the IEEE Conference on Computer Vision and Pattern
  Recognition (2018)

\bibitem{wang2021one}
Wang, T.C., Mallya, A., Liu, M.Y.: One-shot free-view neural talking-head
  synthesis for video conferencing. In: Proceedings of the IEEE/CVF Conference
  on Computer Vision and Pattern Recognition. pp. 10039--10049 (2021)

\bibitem{wang2021gfpgan}
Wang, X., Li, Y., Zhang, H., Shan, Y.: Towards real-world blind face
  restoration with generative facial prior. In: The IEEE Conference on Computer
  Vision and Pattern Recognition (CVPR) (2021)

\bibitem{wang2021hififace}
Wang, Y., Chen, X., Zhu, J., Chu, W., Tai, Y., Wang, C., Li, J., Wu, Y., Huang,
  F., Ji, R.: Hififace: 3d shape and semantic prior guided high fidelity face
  swapping. IJCAI  (2021)

\bibitem{Xu_2022_CVPR}
Xu, Y., Deng, B., Wang, J., Jing, Y., Pan, J., He, S.: High-resolution face
  swapping via latent semantics disentanglement. In: Proceedings of the
  IEEE/CVF Conference on Computer Vision and Pattern Recognition (CVPR). pp.
  7642--7651 (June 2022)

\bibitem{xu2022mobilefaceswap}
Xu, Z., Hong, Z., Ding, C., Zhu, Z., Han, J., Liu, J., Ding, E.:
  Mobilefaceswap: A lightweight framework for video face swapping. AAAI  (2022)

\bibitem{xu2021facecontroller}
Xu, Z., Yu, X., Hong, Z., Zhu, Z., Han, J., Liu, J., Ding, E., Bai, X.:
  Facecontroller: Controllable attribute editing for face in the wild. AAAI
  (2021)

\bibitem{yang2021gan}
Yang, T., Ren, P., Xie, X., Zhang, L.: Gan prior embedded network for blind
  face restoration in the wild. In: Proceedings of the IEEE/CVF Conference on
  Computer Vision and Pattern Recognition. pp. 672--681 (2021)

\bibitem{zhou2021pose}
Zhou, H., Sun, Y., Wu, W., Loy, C.C., Wang, X., Liu, Z.: Pose-controllable
  talking face generation by implicitly modularized audio-visual
  representation. In: Proceedings of the IEEE Conference on Computer Vision and
  Pattern Recognition (CVPR) (2021)

\bibitem{zhu2020aot}
Zhu, H., Fu, C., Wu, Q., Wu, W., Qian, C., He, R.: Aot: Appearance optimal
  transport based identity swapping for forgery detection. Advances in Neural
  Information Processing Systems  (2020)

\bibitem{zhu2021barbershop}
Zhu, P., Abdal, R., Femiani, J., Wonka, P.: Barbershop: Gan-based image
  compositing using segmentation masks. arXiv preprint arXiv:2106.01505  (2021)

\bibitem{Zhu_2021_CVPR}
Zhu, Y., Li, Q., Wang, J., Xu, C.Z., Sun, Z.: One shot face swapping on
  megapixels. In: Proceedings of the IEEE/CVF Conference on Computer Vision and
  Pattern Recognition (CVPR). pp. 4834--4844 (2021)

\bibitem{zi2020wilddeepfake}
Zi, B., Chang, M., Chen, J., Ma, X., Jiang, Y.G.: Wilddeepfake: A challenging
  real-world dataset for deepfake detection. In: Proceedings of the 28th ACM
  International Conference on Multimedia. pp. 2382--2390 (2020)

\end{thebibliography}
